\documentclass[conference]{IEEEtran}
\IEEEoverridecommandlockouts
\usepackage{cite}
\usepackage{amsmath,amssymb,amsfonts}
\usepackage{algorithmic}
\usepackage{graphicx}
\usepackage{textcomp}
\usepackage{xcolor}
\usepackage{hyperref}
\usepackage{cite}
\usepackage{amsmath}
\usepackage{mathtools}
\def\BibTeX{{\rm B\kern-.05em{\sc i\kern-.025em b}\kern-.08em
    T\kern-.1667em\lower.7ex\hbox{E}\kern-.125emX}}
\begin{document}


\title{Multimodal Lengthy Videos Retrieval Framework and Evaluation Metric}


\author{
\IEEEauthorblockN{Mohamed Eltahir, Osamah Sarraj, Mohammed Bremoo, Mohammed Khurd, \\ Abdulrahman Alfrihidi, Taha Alshatiri, and Mohammad Almatrafi}
\IEEEauthorblockA{
King Abdullah University of Science and Technology (KAUST), Thuwal, Saudi Arabia\\
Emails: mohamed.hamid@kaust.edu.sa, \{osamah.sarraj, mohabremoo,
mohamedalawi211\}@gmail.com, \\ \{frihidimany, tahaalshatiri, moalmatrafi90\} @gmail.com
}
\\
\IEEEauthorblockN{Tanveer Hussain}
\IEEEauthorblockA{
Edge Hill University, Ormskirk, England\\
Email: hussaint@edgehill.ac.uk
}
}


\maketitle

\begin{abstract}\\ Precise video retrieval requires multi-modal correlations to handle unseen vocabulary and scenes, becoming more complex for lengthy videos where models must perform effectively without prior training on a specific dataset. We introduce a unified framework that combines a visual matching stream and an aural matching stream with a unique subtitles-based video segmentation approach. Additionally, the aural stream includes a complementary audio-based two-stage retrieval mechanism that enhances performance on long-duration videos. 
Considering the complex nature of retrieval from lengthy videos and its corresponding evaluation, we introduce a new retrieval evaluation method specifically designed for long-video retrieval to support further research. We conducted experiments on the YouCook2 benchmark, showing promising retrieval performance. The framework and evaluation code will be available on GitHub.

\end{abstract}

\begin{IEEEkeywords}
Video moment retrieval, long-form video analysis, multimodal matching, temporal localization, audio filtering.
\end{IEEEkeywords}

\section{Introduction}
In today’s digital world, efficiently accessing information from reliable sources has become increasingly essential. However, the sheer volume of data available continues to grow exponentially, particularly the video content. For instance, on YouTube alone, 500 hours of video content are uploaded every minute, amounting to roughly 82 years of content each day as of 2022 \cite{youtube_stat}. This overwhelming volume of information makes it nearly impossible for individuals to locate relevant information quickly, especially in long-form videos.\

Video retrieval, built on top of deep learning algorithms, addresses this challenge by efficiently locating video segments that match a textual or visual query. Video retrieval has many real-world applications, ranging from security and surveillance to medical videos retrieval.

While many previous works have focused primarily on visual frames and, to a lesser extent, audio transcripts, they often fail to scale to lengthy videos effectively. To address this,  we introduce a multi-modal retrieval framework that adaptively segments videos using subtitles, processes visual and aural information independently, and combines them to retrieve clips more accurately.   

Additionally, to enhance our framework's robustness for lengthy videos, we introduce a two-stage retrieval mechanism within the aural stream. The evaluation of lengthy videos retrieval remains another challenge for researchers, where we introduce a novel evaluation mechanism designed specifically to evaluate querying in long videos.

In recent years, video retrieval has seen a significant increase in demand due to the vast amount of visual content on the Internet. The available systems currently provide multiple methodologies that use technologies such as video-text alignment and multimodal models. Despite their strength and advancement, these technologies still face challenges regarding long video understanding and generality of use zero-shot usage. Zero-shot methods are designed to enable the use of models without requiring additional training from the end user, running without labeled data or extra prompting. 

While short videos retrieval is comparatively easy, lengthy videos carry the context for extended duration, creating challenges for researchers to grab the overall semantics of the retrieved segments. The multimodal short video understanding field has grown significantly thanks to text-image models. For example, VideoCoCa \cite{videococa} uses pre-trained text-image models alongside contrastive and generative technologies to improve text-video retrieval; however, this model focuses on shorter video datasets. Another model that works on short videos is Flamingo \cite{flamingo}, a few-shot model that uses a limited labeled sample for fine-tuning video-text retrieval. However, its reliance on labeled data limits its usability as a zero-shot model while also struggling with complex temporal dependencies, that exist in many real-world applications including endoscopy videos retrieval in medical domain. \\

A crucial aspect of our work is the focus on zero-shot usability. Norton \cite{norton}, a model that utilizes multimodal pre-training and contrastive learning, can achieve excellent results on short and medium-length videos. However, its performance on longer videos with zero-shot applications falls short of the desired performance.\\


Recent advances in vision-language models have shown promising results in video understanding. A recent study \cite{clip4clip} introduces CLIP4Clip, which adapts CLIP's image-text matching capabilities for video by incorporating temporal information through similarity calculation and mean pooling, achieving strong performance on video retrieval tasks. Building on this, \cite{videoclipxl} presents VideoCLIP-XL, a larger-scale model that enhances video-text alignment through contrastive learning and temporal modeling, demonstrating improved performance on various video understanding benchmarks. Additionally, \cite{internvid} introduces InternVid and ViCLIP, which utilize a multi-stage training approach and a novel video-text alignment strategy. ViCLIP, in particular, stands out for its efficient handling of temporal dependencies and robust performance on diverse video datasets. However, these models still face challenges with longer videos and complex temporal relationships, areas where our approach offers significant improvements.

\begin{figure*}[htbp]
\centerline{\includegraphics[width=\textwidth]{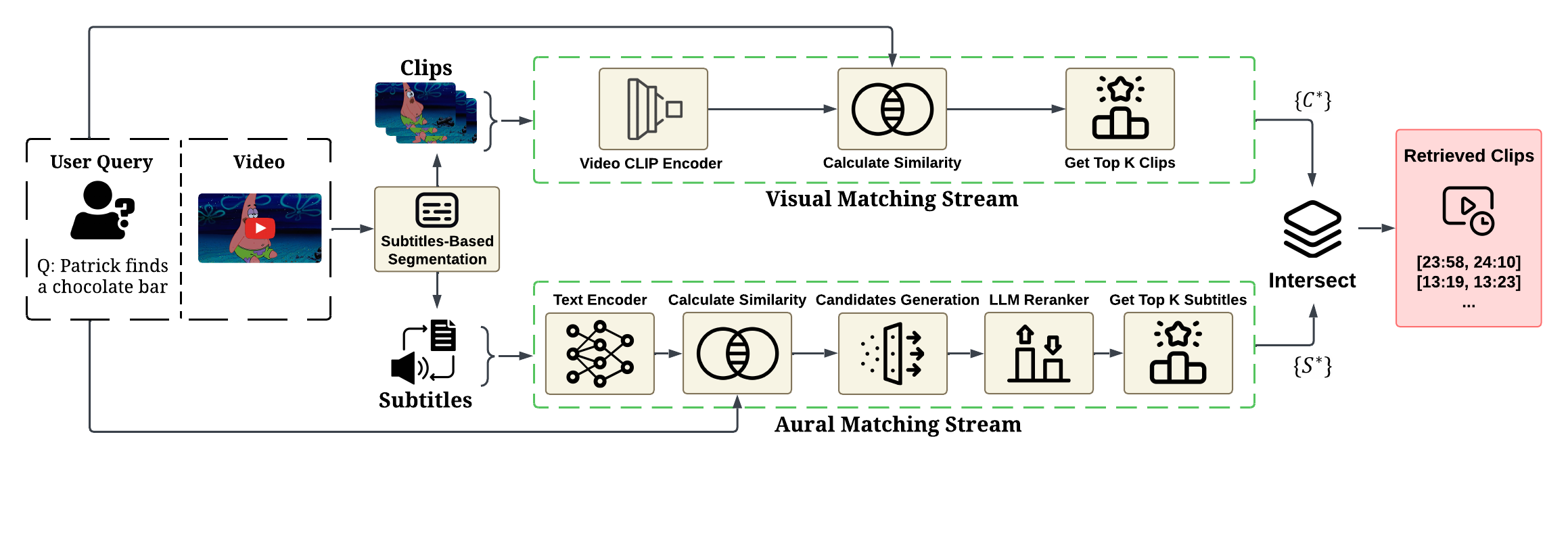}}
\vspace{-30pt}
\caption{The proposed framework for multi-modal video querying. The visual stream retrieves the top $K$ clips ($\textit{C*}$), while the aural stream retrieves the top $K$ subtitles ($\textit{S*}$). The final retrieved clips are obtained by intersecting $\textit{C*}$ and $\textit{S*}$, ranked by averaged similarity scores.}
\label{fig_pipeline_1}
\end{figure*}

\section{The Proposed Method}
The proposed framework is multi-modal video querying architecture introduced to provide high-level adaptability and scalability for different video understanding tasks. It utilizes multi-modal data by independently processing the aural and visual information of the input video at the beginning, then align them at the final stage to retrieve the most relevant clips. Moreover, our framework removes the dependency on prior segmentation of the input video, which is common in existing video retrieval techniques. Additionally, to enhance our framework robustness for lengthy videos, we introduce a two-stage retrieval mechanism within the aural stream, that is, consists of a candidates generator and a re-ranker. Finally, we evaluate the proposed framework using our novel evaluation mechanism designed specifically to evaluate querying in long videos. The key concepts are explained in detail in the following sections.

\noindent\textbf{Inputs Preprocessing:}
Initially, we acquire an input video, extract and transcribe the audio data using a pre-trained speech recognition model. The key idea is not only to extract subtitles from the audio; rather, we are interested in utilizing the start and end timestamps provided by the model, along with the subtitles, to segment the video into clips ($\textit{C}$). Unlike fixed-interval segmentation, this Subtilites-based Segmentation in our framework is unique and also more robust as it aligns with the natural pauses of speech, ensuring the overall consistency of each clip.

\subsection{Visual Data Processing}
In the visual matching stream, each clip ($\textit{C}_i$) acquired from the preprocessing stage along with the user query ($\textit{Q}_{user}$) is processed using a pre-trained embedding model ($\phi_{\textit{CLIP}}$) to create clip embeddings $\phi_{\textit{CLIP}}(\textit{C}_i)$ and query embeddings $\phi_{\textit{CLIP}}(\textit{Q}_{\textit{user}})$, denoted as ($\phi_{\textit{CLIP}, \textit{C}_i}$) and ($\phi_{\textit{CLIP}, \textit{Q}}$), respectively. The $\phi_{\textit{CLIP}}$ is a hybrid model designed to jointly process and align video and text modalities in a common embedding space, based on Contrastive Language-Image Pre-Training (CLIP) model, and is commonly referred to as video CLIP. 
This video CLIP model $\phi_{\textit{CLIP}}$ aligns the generated representations ($\phi_{\textit{CLIP}, \textit{C}}, \phi_{\textit{CLIP}, \textit{Q}}$) within a common embedding space. Finally, the cosine similarity metric, which measures the angular similarity between embeddings, is used to find the relevancy between $\phi_{\textit{CLIP}, \textit{C}}$ and $\phi_{\textit{CLIP}, \textit{Q}}$. The cosine similarity is defined as:

\[
\text{Cosine Similarity}(\phi_{\textit{CLIP}, \textit{C}_i}, \phi_{\textit{CLIP}, \textit{Q}}) = \frac{\phi_{\textit{CLIP}, \textit{C}_i} \cdot \phi_{\textit{CLIP}, \textit{Q}}}{\|\phi_{\textit{CLIP}, \textit{C}_i}\| \|\phi_{\textit{CLIP}, \textit{Q}}\|}
\]

Considering the similarity between clips and query embeddings, we retrieve the top K clips ($\textit{C*}$), where $\textit{C*}$ is a subset of $\textit{C}$.

\subsection{Aural Data Processing}
As for the aural processing stream, we encode the clips subtitles ($\textit{S}$) extracted from the pre-trained speech recognition model and the user query ($\textit{Q}_{\textit{user}}$) using a text encoder ($\phi_{\textit{text}}$). This generates subtitle embeddings $\phi_{\textit{text}}(\textit{S})$ and query embeddings $\phi_{\textit{text}}(\textit{Q}_{\textit{user}})$, denoted as ($\phi_{\textit{text}, \textit{S}}$) and ($\phi_{\textit{text}, \textit{Q}}$), respectively.
The cosine similarity is then computed as:
\[
\text{Cosine Similarity}(\phi_{\textit{text}, \textit{S}}, \phi_{\textit{text}, \textit{Q}}) = \frac{\phi_{\textit{text}, \textit{S}} \cdot \phi_{\textit{text}, \textit{Q}}}{\|\phi_{\textit{text}, \textit{S}}\| \|\phi_{\textit{text}, \textit{Q}}\|}
\]
Using this similarity measure with embeddings, we capture subtitles that are semantically similar to the user query. Subsequently, we retrieve the top K semantically similar subtitles list.
This list is fed into the candidate generation component, which enriches the list with additional samples obtained through a single heuristic: including all subtitles containing at least one occurrence of a word from the user query. This particularly helps capture samples that are lexically similar, meaning those containing exact matches or partial overlaps with the query words. Note that while only one heuristic was applied in our framework, it could be extended with more rules to improve the retrieval of relevant samples that were not captured by the semantic similarity measure. Therefore, in this setup, the text encoder focuses on capturing semantically similar samples, while the heuristic focuses on lexically relevant ones. These are then combined to produce an extended top K subtitles list. This list is subsequently passed to a Large Language Model ($\phi_{\textit{LLM}}$) along with the user query ($\textit{Q}_{\textit{user}}$) to be re-ranked based on their relevance, resulting in the final ranked top K subtitles ($\textit{S}^*$). 

\subsection{Aural-Visual Alignment via Intersection}
Given video segmentation, based on subtitles' timestamps, is identical across both streams, the clips and subtitles share the same unique timestamp intervals. We align the top K clips and the top K subtitles by selecting the clips that share the same timestamp intervals in the top K results ($\textit{C*}$ and $\textit{S*}$), that is, intersecting the top k of the two streams. Subsequently, we compute the average of the cosine similarities from both the top K aural and visual streams samples and rank the results based on these averaged similarity scores to obtain the final retrieved clips.

It is important to note that even the lexically retrieved samples have similarity scores, as all embeddings were calculated by the text encoder before the candidate generation step.

This alignment approach is unique as it aligns the clips directly at the final stage by intersecting the results from both streams based on timestamp intervals, whereas most previous works align the embeddings (i.e., intermediate representation) internally during processing. This alignment approach is made possible because the segmentation, based on shared timestamps, at the beginning is consistent across both streams, allowing us to perform an intersection at the end. Additionally, our method does not depend on prior segmentation of the input video, enabling it to segment any video adaptively. This makes it adaptable to new, unseen audible videos. Furthermore, the proposed two-stage retrieval mechanism for the aural stream gives our framework the ability to handle lengthy videos with high retrieval accuracy.

\section{Experiments and Results}
\noindent\textbf{Implementation Details:}
For our framework, we utilized Whisper Large V3 Turbo \cite{whisper}, an optimized variant of Whisper Large V3, as our pre-trained speech recognition model to segment the input video and extract timestamps and subtitles. Whisper is widely recognized as a state-of-the-art (SOTA) speech recognition model. For the video CLIP model ($\phi_{\textit{CLIP}}$), we experimented with four variants, including CLIP4Clip \cite{clip4clip}, ViCLIP-base, ViCLIP-large \cite{internvid}, and VideoCLIP-XL \cite{videoclipxl}, which have demonstrated strong performance in video understanding tasks, as highlighted in VideoCLIP-XL paper. For the text encoder in the aural stream ($\phi_{\textit{text}}$), we utilized the Stella 400M \cite{stella} embeddings model, which ranks highly on the Massive Text Embedding Benchmark (MTEB) leaderboard \cite{mteb}, demonstrating its robustness in text encoding tasks. In our setup, we set $K=30$ for the top K samples output by the text encoder, which are passed to the candidate generation component. The number of candidates can vary depending on the heuristic used to extend the list. For the re-ranking model ($\phi_{\textit{LLM}}$), we used Gemini 2.0 Flash \cite{gemini}, a SOTA large language model developed by Google, to re-rank the retrieved subtitles based on their relevance to the user query. At the final stage, we select the top $K=10$ samples for each stream ($\textit{S}^*$ for the aural stream and $\textit{C}^*$ for the visual stream). All experiments were run on a single NVIDIA A100-80GB GPU.

The benchmarking of the proposed method is challenging because current evaluation methods are primarily designed for retrieving short clips from highly diverse videos, which are not well-suited for our approach that focuses on lengthy single audible videos, such as movies or lectures. To address this limitation, we propose a new evaluation method specifically designed for long audible videos retrieval.

The evaluation method starts by taking the timestamp intervals of the top K clips predicted by the model and intersecting them with the ground truth intervals provided by the benchmark dataset for the input video. A prediction is considered a match if the intersection exceeds a predefined threshold (Intersection Threshold). Recall metrics are then calculated based on these matches, providing a performance measure for the model's ability to retrieve relevant intervals within the same video.

We investigated several benchmark datasets, including MSRVTT \cite{msrvtt}, DiDeMo \cite{didemo}, LSMDC \cite{lsmdc}, and MSVD \cite{msvd}, but found them unsuitable for our approach due to their short clip durations or lack of audio content. To evaluate our method effectively, we focused on datasets containing long audible videos, allowing for independent evaluation within each video.

Consequently, we selected YouCook2 \cite{youcook2} to benchmark our work. This dataset consists of 2,000 untrimmed instructional cooking videos, with a total duration of 176 hours and 5.27 minutes on average per video. Following the splits defined by \cite{howto100m}, we have 3,305 test clip-text pairs from 430 videos for zero-shot retrieval evaluation. All videos include audio, making YouCook2 well-suited for our evaluation. The clips intervals were used as ground truth, while their corresponding text used as the input query. Unlike traditional approaches, which use predefined clips for evaluation and perform retrieval across the clips of the entire dataset, our framework processes the full videos directly, handles the splitting dynamically, performs retrieval over the clips of each video independently and uses the predefined clips only as ground truth for evaluation.

To evaluate the framework performance, we used Average Recall@K, which measures the proportion of relevant samples retrieved in the top K results, across intersection thresholds from 0.5 to 0.95, with a step size of 0.05. Intersection threshold represents the temporal overlap between retrieved and ground truth segments. This averaged metric is similar to the concept of mean Average Precision (mAP) @ K in object detection, where metrics are averaged over a range of intersection thresholds from 0.5 to 0.95, with a step size of 0.05, to assess model's performance across varying levels of overlap. 

As shown in \tableautorefname{ 1}, employing VideoCLIP-XL within the proposed framework outperforms all the other video CLIP models across all metrics, especially at higher recall thresholds. This shows its superiority in capturing vision-text relationships compared to the other models. 
Figures 2, 3 and 4 demonstrate the performance change across different intersection thresholds. We can notice how recall scores decrease as the intersection threshold value increases, indicating that stricter overlap requirements reduce the number of relevant matches.
Furthermore, Figures 5 and 6 show qualitative examples that provide further insights into the performance gap between these video CLIP models. These samples highlight VideoCLIP-XL's consistent ability to retrieve clips accurately at higher ranks compared to the others.

However, an interesting insight that can be seen is that the benchmark itself presents significant challenges for the models. One might assume that reducing the search space (i.e., focusing on fewer clips from a single long video rather than hundreds of clips across diverse multiple videos) would allow the models to achieve near-perfect scores effortlessly. But contrary to this expectation, the results reveal otherwise. The highest Average Recall@1 score is only 28.5\%, highlighting a significant opportunity for improvements in these models and in the framework in general.


\begin{table}[htbp]
    \caption{Average Recall@K scores for zero-shot text-video retrieval on YouCook2 benchmark using the proposed framework. R stands for Recall.}
    \begin{center}
        \begin{tabular}{l|c|c|c}
        \hline
            \textbf{Video CLIP Model} & \textbf{Avg R@1} & \textbf{Avg R@5} & \textbf{Avg R@10} \\
        \hline
            CLIP4Clip \cite{clip4clip} & 27.03\% & 40.68\% & 41.09\% \\
            ViCLIP-B \cite{internvid} & 24.46\% & 36.09\% & 36.35\% \\
            ViCLIP-L \cite{internvid} & 28.52\% & 42.62\% & 42.89\% \\
            VideoCLIP-XL \cite{videoclipxl} & \textbf{28.56}\% & \textbf{44.13}\% & \textbf{44.41}\% \\
        \hline
        \end{tabular}
    \label{tab_results}
    \end{center}
\end{table}

\vspace{10cm}

\begin{figure}[htbp]
\centerline{\includegraphics[width=\columnwidth]{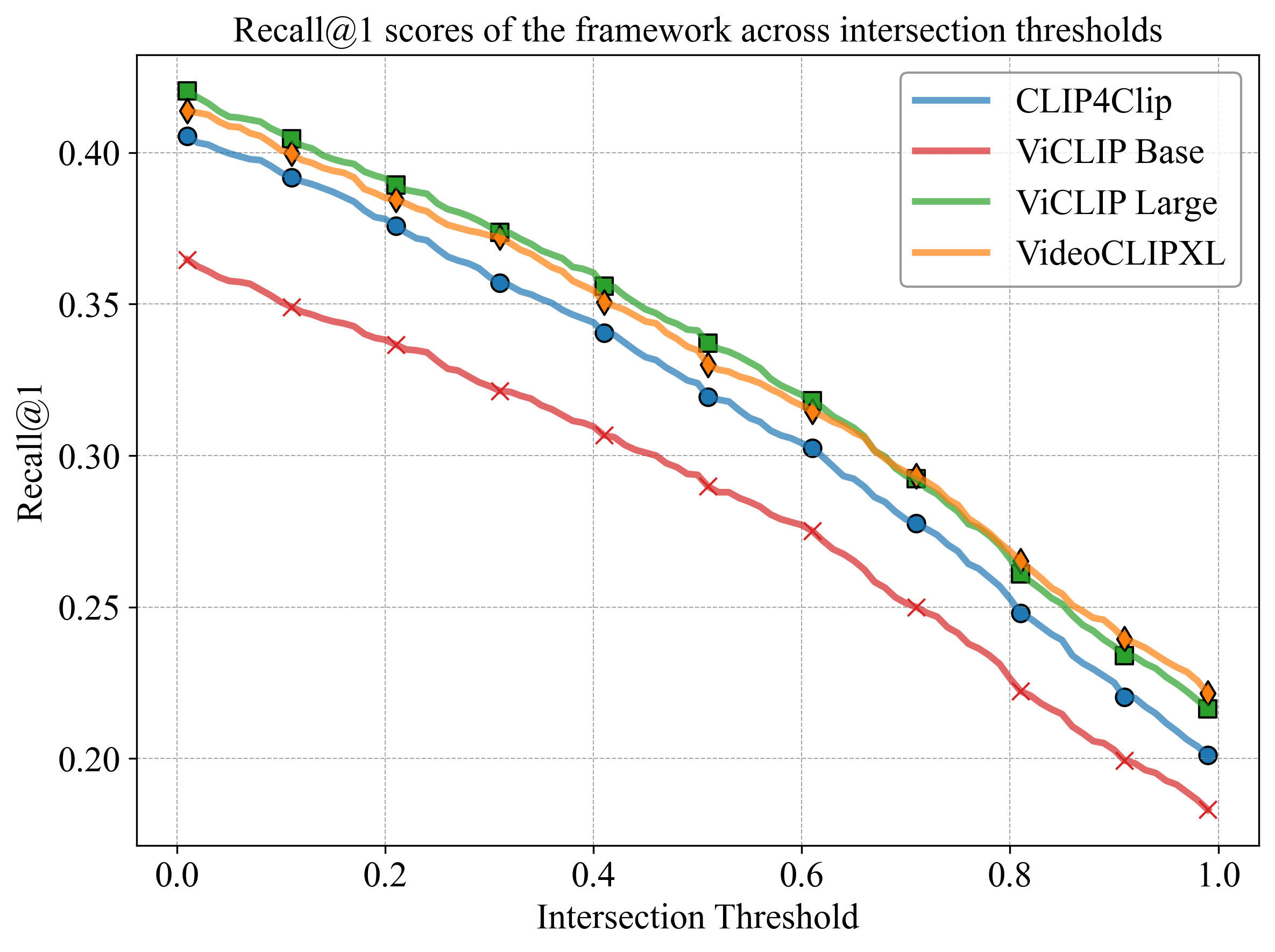}}
\caption{Recall@1 scores of the framework across intersection thresholds.}
\label{fig_pipeline_2}
\end{figure}

\begin{figure}[htbp]
\centerline{\includegraphics[width=\columnwidth]{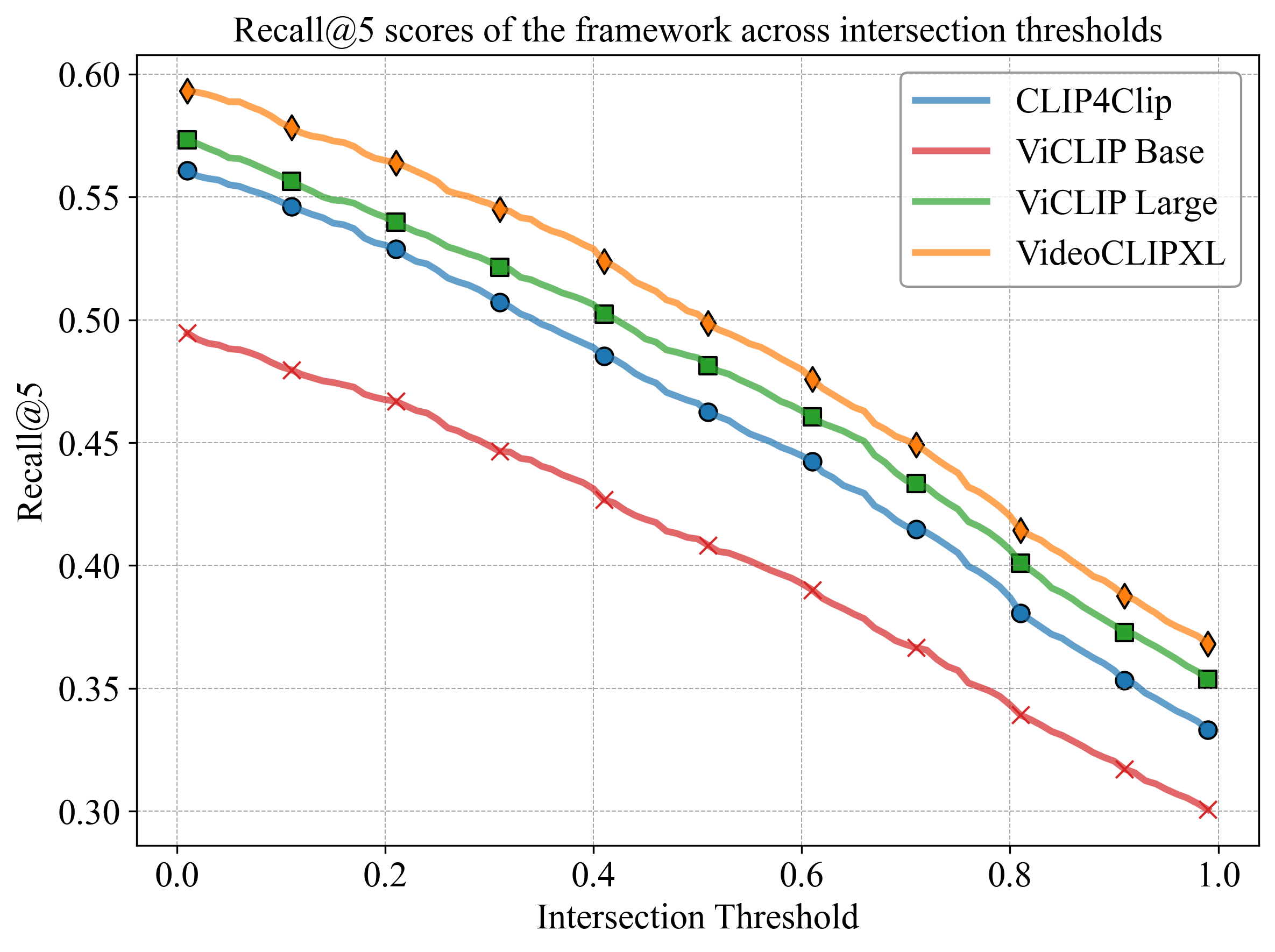}}
\caption{Recall@5 scores of the framework across intersection thresholds.}
\label{fig_pipeline_3}
\end{figure}

\begin{figure}[htbp]
\centerline{\includegraphics[width=\columnwidth]{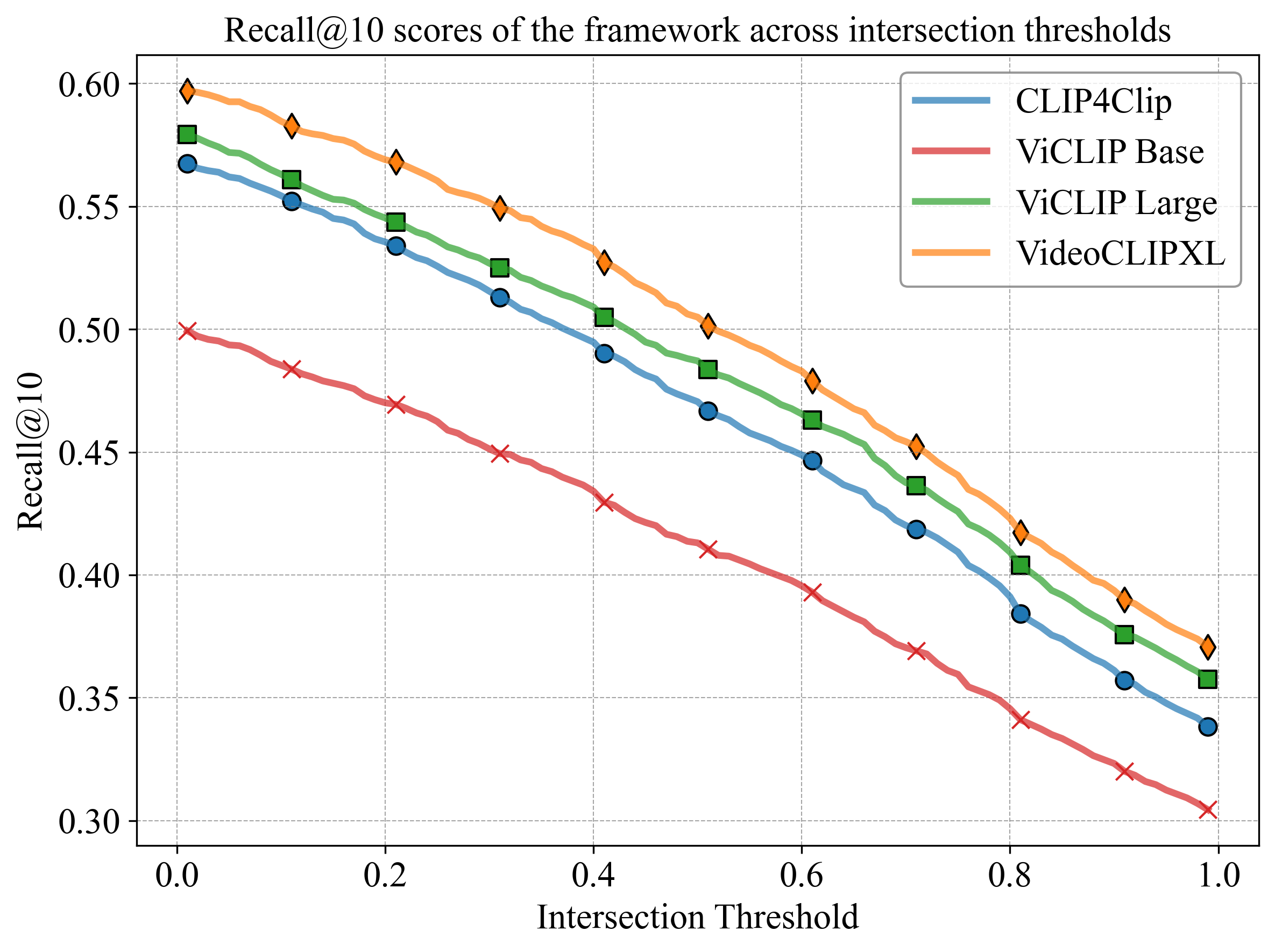}}
\caption{Recall@10 scores of the framework across intersection thresholds.}
\label{fig_pipeline_4}
\end{figure}

\begin{figure*}[htbp]
\fbox{\centerline{\includegraphics[width=\textwidth]{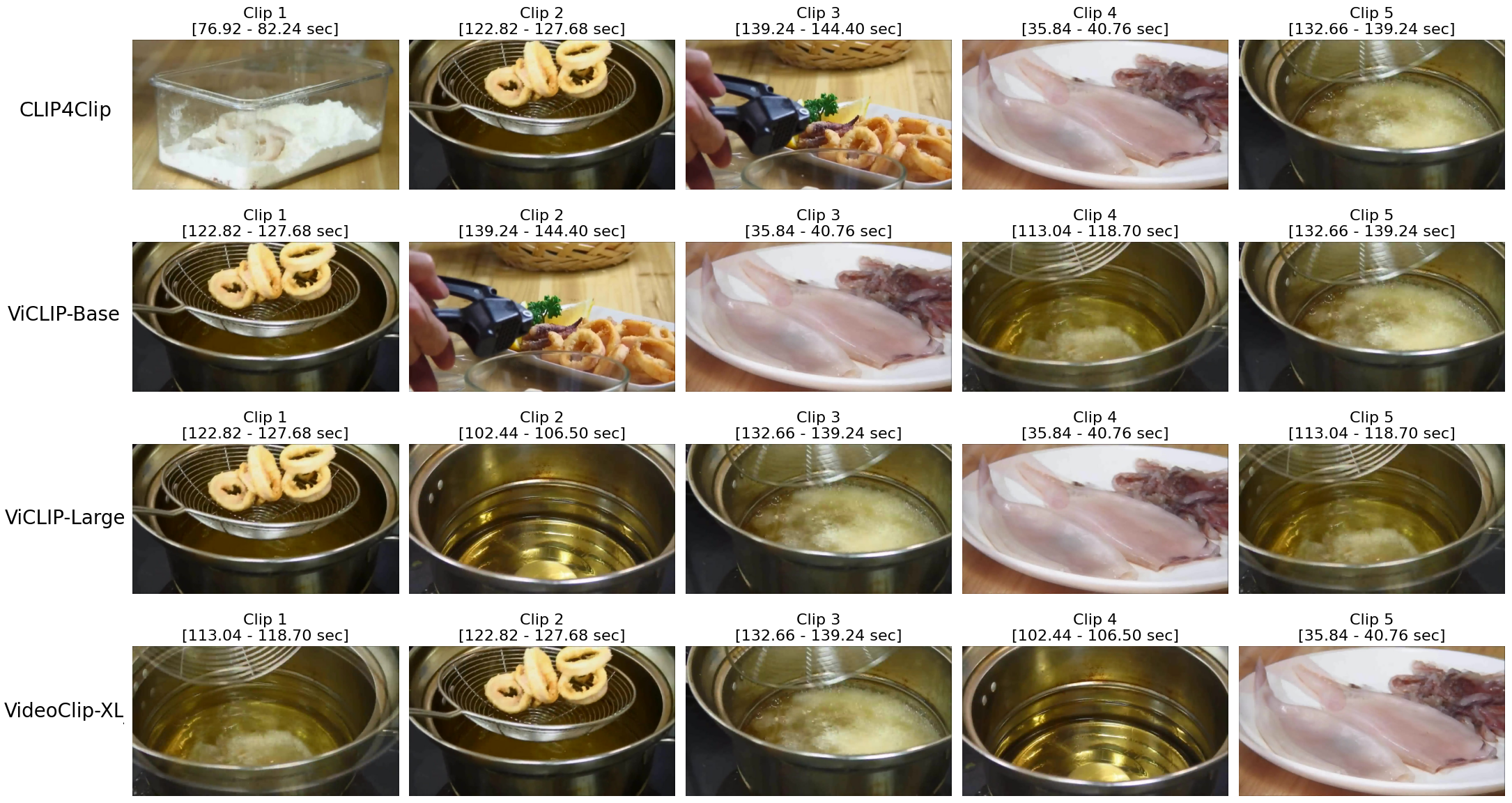}}}
\caption{Qualitative results for the query: "Add the squid into a pot of hot oil". The ground truth interval is [114, 121]. CLIP4Clip failed to retrieve the correct interval within the top 5 clips. ViCLIP, with its base and large variants, retrieved the correct interval in the 4th and 5th positions, respectively. VideoCLIP-XL managed to retrieve the correct clip as the first result, demonstrating its powerful retrieval performance.}
\label{fig_pipeline_5}
\end{figure*}

\begin{figure*}[htbp]
\fbox{\centerline{\includegraphics[width=\textwidth]{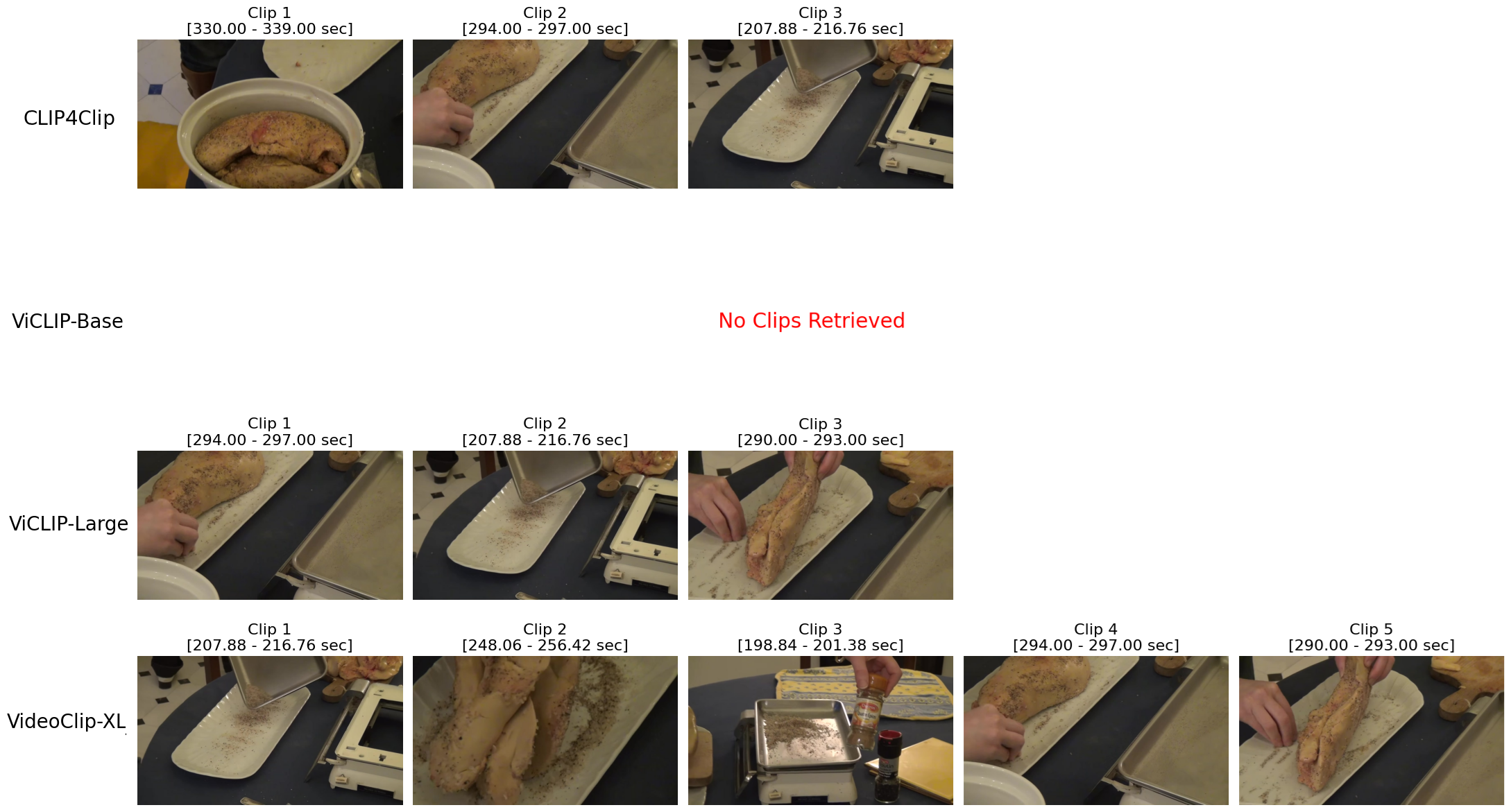}}}
\caption{Qualitative results for the query: "Sprinkle the marinade powder on a plate". The ground truth interval is [201, 220]. The results show the behavior of the framework when intersecting the top K clips and subtitles. ViCLIP-Base failed to retrieve any relevant matches, indicating its inability to align the top clips and subtitles. On the other hand, VideoCLIP-XL successfully retrieved the correct clip as its first result, while the other models (ViCLIP-Large and CLIP4Clip) retrieved it in their second and third ranks, respectively.}
\label{fig_pipeline_6}
\end{figure*}



\vspace{500cm}
\vspace{500cm}
\vspace{500cm}
\vspace{500cm}
\vspace{500cm}

\section{Conclusions and Limitations}
We proposed a multi-modal video querying framework that processes aural and visual streams independently and aligns them at the final stage. By segmenting videos using subtitle timestamps, the framework removes dependency over predefined clips, making it adaptable to unseen videos. Our evaluation on the YouCook2 dataset showed the effectiveness of this approach, with VideoCLIP-XL outperforming other video CLIP models across the retrieval metrics.

Although the subtitles-based segmentation approach provides consistent clips, a key limitation is that it generates only short clips, which may fail to retrieve the correct segment when the desired clip is longer.

Future work could work on addressing this limitation alongside improving the framework by incorporating more robust heuristics for candidates generation and utilizing better encoders and re-rankers. It could also work on extending the framework applicability to diverse datasets and video understanding tasks.

\bibliographystyle{IEEEtran}

\end{document}